\documentclass[a4paper]{article}
\pdfoutput=1
\usepackage{hyperref}
\hypersetup{
  pdfinfo={
    Title={YA framework for mining process models from email logs},
    Author={Diana Jlailaty},
    Subject={...},
    Keywords={...}
  }
}
\usepackage{tikz}
\usepackage{amsmath}
\usepackage{color,xparse,l3regex,hyperref}
\usepackage{chngcntr}
\usepackage[english]{babel}
\usepackage[T1]{fontenc}
\usepackage[utf8]{inputenc}
\usepackage{amsmath}
\usepackage{graphicx}
\usepackage{amssymb}
\usepackage{mathtools}
\usepackage{csvsimple}
\usepackage{placeins}
\usepackage{authblk}
\usepackage[colorinlistoftodos]{todonotes}
\usepackage{xpatch}
\usepackage{url}
\usepackage{pdfpages}
\usepackage[compact]{titlesec}
\titlespacing{\section}{0pt}{*0}{*0}
\titlespacing{\subsection}{0pt}{*0}{*0}
\titlespacing{\subsubsection}{0pt}{*0}{*0}
\counterwithout{table}{section}

\usepackage{pdfpages}
\begin{document}

\title{A framework for mining process models from email logs}

\author[1]{Diana Jlailaty}
\author[1]{Daniela Grigori}
\author[1]{Khalid Belhajjame}
\affil[1]{Paris Dauphine University, PSL Research University, CNRS, UMR[7243], LAMSADE, 75016 PARIS, FRANCE}


\maketitle              



\begin{abstract}
Due to its wide use in personal, but most importantly, professional contexts, email represents a valuable source of information that can be harvested for understanding, reengineering and repurposing undocumented business processes of companies and institutions. 
Towards this aim, a few researchers investigated the problem of extracting process oriented information from email logs in order to take benefit of the many available process mining techniques and tools. In this paper we go further in this direction, by proposing a new method for mining process models from email logs that leverage unsupervised machine learning techniques with little human involvement. Moreover, our method allows to semi-automatically label emails with activity names, that can be used for activity recognition in new incoming emails. A use case demonstrates the usefulness of the proposed solution using a modest in size, yet real-world, dataset containing emails that belong to two different process models.

\end{abstract}

\section{Introduction}
Email is by and large the most popular communication medium, and is considered by some the first and largest social media \footnote{ http://www.web-strategist.com/blog/2009/07/09/email-the-first-social-network/}.  While its initial use focused on exchanging (personal) messages between individuals, email is nowadays used for complex activities ranging from organization of events, to sharing and editing documents, to coordinating the execution of complex tasks involving multiple individuals. A recent study has shown that email is still the primary method of communication, collaboration and information sharing\footnote{ http://onlinegroups.net/blog/2014/03/06/use-email-for-collaboration/ }.

Because of its wide use in personal, but most importantly, professional contexts, email represents a valuable source of information that can be harvested for understanding, reengineering and repurposing undocumented business processes of companies and institutions. Towards this aim, a few researchers investigated the problem of mining processes from email logs. For example, Aalst et al. \cite{Aalst1-2007} developed EmailAnalyzer, a tools for transforming  email messages in MS Outlook into a format that can be used by process mining tools. Mavaddat et al. \cite{Mavaddat-2011} elaborated an approach to classify emails into business and non business oriented categories. Many approaches propose to label emails with speech acts (see for example, \cite{Khoussainov-2006}).
While useful, existing proposals use mainly supervised machine learning techniques, thereby requiring human-supplied training datasets. Moreover, they sometimes make assumption that do not hold always in practice. For example, the work by \cite{Aalst1-2007} assumes that the email message subject includes the activity name.\\\\We present in this paper a new method for mining process models from email logs that leverage unsupervised machine learning techniques with little human involvement to map email logs into process models. We do not assume that emails have been (manually) preprocessed. Instead, we start with raw emails that undergo a number of (automatic) preprocessing steps for cleansing them and translating them into a format that is expected by process mining tools.\\

Given that advanced process mining techniques and tools exist, the challenge we face is to extract from emails the information required by this techniques:
\begin{itemize}
\item among the many business processes in which the user is involved, which one is concerned by a given email? 
\item which process activity is related to the given email?
\item given a process model, what are the emails that belong to the same process instance? 
\end{itemize}

To address this challenge, we adopt an iterative approach that given a log of emails, clusters preprocessed emails to identify process models and instances, then uses knowledge about process instances to guide activity recognition.\\

The contributions of the paper are therefore as follows:
\begin{itemize}
\item
An approach for mining the process models underlying a log of emails. It is worth noting here that the emails used as input may belong to different process models, and that our approach is able to cope with this kind of heterogeneity.  Process related information is exctracted from emails, allowing to benefit from existing process mining tools. 
\item
An implementation of the steps of the approach using state of the art techniques, such as hierarchical and k-means clustering, customized for our needs. 
\item 
A method to semi-automatically label emails with activity names, with little user effort, instead of manually labeling the emails in the training set as done by speech act based approaches. The labelled emails can be used for activity recognition in new incoming emails and thus for recommendations about adding activities in user task list. 
\item
A use case that demonstrates the usefulness of the proposed solution using a modest in size, yet real-world, dataset containing emails that belong to two different process models, namely meeting organization and requests for conference travel grants.  
\end{itemize}

The paper is organized as follows. We start by overviewing our approach in Section \ref{sec:approach}. 
We present in detail the phases that compose our approach in Sections \ref{sec:phase1},\ref{sec:phase2} and\ref{sec:phase3}. We then present a use case that showcase our approach using a dataset of real-world emails in Section \ref{sec:usecase}. An analysis of related works is presented in \ref{sec:relatedwork} before concluding the paper in Section \ref{sec:conclusions}. 
\section{Approach Overview}
\label{sec:approach}
\begin{figure}[ht]
 \includegraphics[width=\linewidth]{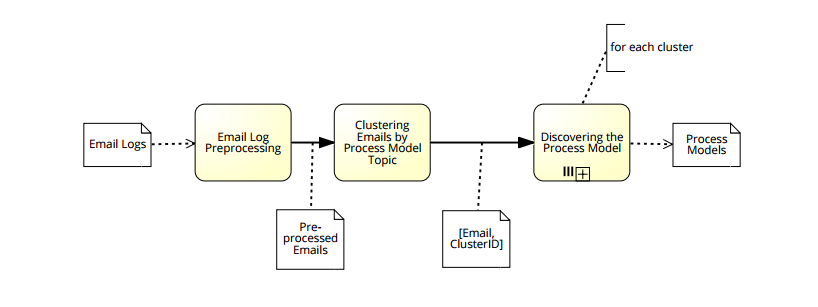}
  \caption{Approach overall steps (using BPMN notation)}
  \label{fig:approachSteps}
\end{figure}

Figure \ref{fig:approachSteps} shows that our approach is a three-step process that given a set of emails as input, produces a set of process models. The emails can be supplied by an individual, for example, a student or a researcher, or an institution.

\FloatBarrier
\FloatBarrier
In order to be more clear in the rest of the paper, we will provide some terms definitions:
\begin{enumerate}
\item {\bf Process Model}: is the generic representation of all process instances.Process model describes a business process that is applied for achieving specific goal. It is a connected graph of nodes and edges. Nodes represent the activities applied during the process enactment and the edges connecting the nodes represent the flow between the activities. Mainly, process mining techniques are applied on event logs to derive their corresponding process models.
\item {\bf Instance/Case}: An instances is a specific execution of process model. There may be several ways for executing the same business process. This will lead to the formation of various process instances originating from the same model. An instance is composed of a set of connected activities.
\item {\bf Activity}: is the core of a process model. An activity is a task that should be accomplished in defined period of time to work towards the process goals. It can be broken down into a set of timestamped events.
\item {\bf Process Topic}: When using this expression, we mean the topic or the subject the process revolving around such as meeting scheduling, applying for Ph.D position, funding for conference attendance, requesting documents.. etc
\end{enumerate}
In the first step, the emails provided as input are preprocessed in order to cleanse and translate them into a structured format that is amenable for mining. The contents of the resulting structured data is then normalized. The resulting data undergo a clustering phase in which emails are group. At the end of this step, each email is assigned an ID, that identifies the cluster to which the email belongs. For each cluster of emails, a process model discovery is applied (step 3). 

\FloatBarrier
\begin{figure}[ht]
 \includegraphics[width=\linewidth]{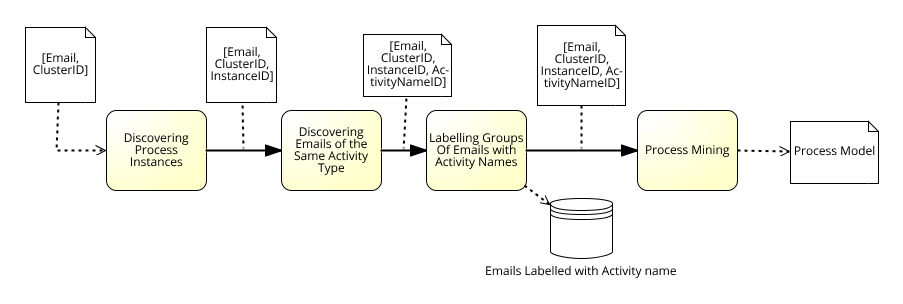}
  \caption{Process model discovery phase}
  \label{fig:processmodeling}
\end{figure}

The third step is composite in the sense that it can be broken down into sub-steps as illustrated in Figure \ref{fig:processmodeling}. Given a cluster of emails that belong to the same process topic, such emails are first sub-grouped into clusters, each representing a process instance (step 1 in Figure \ref{fig:processmodeling}).

The emails of each process instance are then labeled with the type of the activity they belong to. Notice here that multiple emails of the same process instance may belong to the same activity. The second step in Figure \ref{fig:processmodeling} is used for this purpose to cluster emails that belong to the same activity; the third step will associate them with an activity name (based on the labels that the user assigns to the medoids of the clusters). The resulting labeled set is stored and can be used for activity recognition in incoming emails. Finally, an existing process mining technique is utilized to infer the process model. See \cite{Aalst-2004} for a survey on such techniques.\\\\

\section {Phase 1: Email Log Preprocessing}
\label{sec:phase1}
The input of the first step is a set of emails that will be preprocessed in order to cleanse and translate them into a structured format that is amenable for mining.

Emails are unstructured data that should be translated into the format expected by the mining tools.
Some content of the message may be useless for our analysis. For that reason, cleansing of data is an important step.  Moreover, attributes describing emails (used in similarity measurement) can be of different data types. Thus, it is essential to apply some pre-processing and normalization of data before starting the analysis. In the following we will describe the steps of cleansing, representation and normalization of data.

\subsection{Cleansing}
Cleansing is considered an essential step in emails' preprocessing as it eliminates all the elements that may affect badly the clustering quality. We use the text mining library of R statistical language for preprocessing of the input emails. Emails may contain a lot of words or symbols that do not serve our analysis such as the stopwords, numbers, or punctuations. R statistical language provides multiple built-in functions that can perform specific text pre-processing tasks such as removing stopwords, removing numbers, removing punctuation and lowering capital letters etc... This helps in optimizing the similarity measurement between the contents of emails. 
\subsection{Representation of Data} The input data is a csv file in which each row represents an email. The columns represent the attributes describing the email such as sender, receiver, subject, body, timestamp etc...
In this step, our goal is to reformulate the input data to be compatible with the data required by the analysis step of our framework. When data is imported, we can exclude the attributes that are out of our interest. In our case, we import the subject, body, sender, receivers and the timestamp of each e-mail. The imported bodies of emails will be converted into a matrix where the rows are the input emails and the columns are all the words contained in all the emails' bodies. Each entry $e_{ij}$ of this matrix will represent the number of occurrences of each word $w_{j}$ in email $m_{i}$.
\subsection{Data Normalization and Preparation} \label{sub:normalization}
In this step, we normalize and process the email text (the values of the words matrix) in order to improve the clustering results.

Since the number of occurrences of a word in an email may not hold all the needed information for clustering, the built-in TF-IDF method \cite{Ramos-2003} is used. TF short for term frequency and IDF, short for term inverse document frequency. TF-IDF is a numerical statistic that is intended to reflect how important a word is to a document in a collection or corpus..

In our case, we want to summarize a text using the words it includes. One solution is to pick the most frequently occurring words. However, some meaningless words may occur frequently i.e have high term frequency TF value across all documents  such as "this", "a". Hence, it is important to study how infrequently a word occurs across all documents. Therefore, the product of TF-IDF of a word gives a product of how frequent this word is in the document multiplied by how unique the word is w.r.t. the entire corpus of documents.

In our customized distance calculations through out our whole approach, we may include different attributes such as email subject, email body or timestamp. When we say "customized", we mean the method we follow to calculate the distance between documents. In other words, we mean the attributes that we include or the distance functions we use. Knowing that each attribute has its own data type (subject and body are both strings, timestamp is date and time), normalization is considered an important step in the preprocessing phase. We normalize data in a way that all values will be in [0,1] range (for the subject and body attributes). Although the body and the subject attributes are of the same data type, the number of words in the subject is much more smaller than that in the body of the emails. In order to account for texts of different lengths, each TF-IDf vector is normalized so that there is no more differece in the length of the processed texts.

For the timestamp attribute, we set the distances between the timestamps such that the distance is directly proportional to the time difference. In other words, as the duration between two emails increases, the dissimilarity value given will increase as well. This is intuitive, because two emails are less possible to be related if they are sent in very far timings which will be translated in the increase of the distance between two emails. The distance provided is also in the [0,1] interval. 
\\\\

\begin{figure}[ht]
 \includegraphics[width=\linewidth]{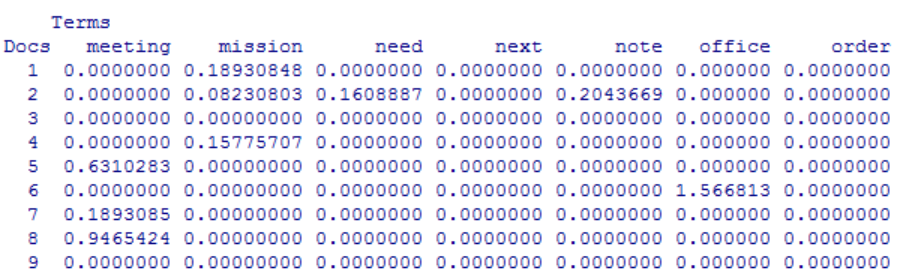}
 \label{fig:sampledata}
  \caption{Sample of the input data}
  \label{fig:inputData}
\end{figure}

Figure \ref{fig:inputData} shows a sample of the body words matrix in which the values of the entries are the normalized TF-IDF of each word in each document.\\\\
\FloatBarrier
\FloatBarrier
TF/IDF score may be not sufficient in order to distinguish the important words.Though we do not depend on the subject of the email to specify its process topic, for our application we consider that words appearing in emails subjects are in a way or another important, they give clues about the main topic of the body. For this reason, we form a set of terms obtained from the words appearing in all emails' subjects. We then traverse the body words matrix and multiply the entries of the words occurring in the obtained word set by a specific weight. The optimum weight is deduced through experimental sensitivity analysis.

\section{Phase 2 - Clustering emails belonging to the same process model}\label{sec:phase2}
In this step, our objective is to group the emails into clusters according to their process topic. 
We can say that all emails in one cluster had been exchanged for achieving a specific goal. So the emails exchanges happening in one cluster between different entities aim for achieving specific activities that correspond to a process model (to be discovered). For example, if we fetch the emails of a researcher, we can see that there are multiple email sets for scheduling a meeting or for organizing a conference mission etc..\\\\
The input of this step is a set of emails taken from a specific user. The emails are acquired from the inbox and sent mails folders. We process the emails separately without having any knowledge about the threading relations between them. A thread starts with an original email which is followed by a sub-sequence of replies between peers allowing them to keep track of past conversations organized in chronological order. We deal with emails separately for three main reasons: (1) some email management  systems do not collect emails in the form of threads (they do not support threading, an example on that is the thunderbird email management system) (\cite{EreraC08}) . (2) In the case where email management systems adopt threading, some threads may start with a specific topic(process) and drifts to another one in later emails. The drift may occur in the content of the message, while the subject is kept the same in the whole thread.(3) In addition, one process may span multiple threads. As an example of this, suppose a Ph.D student is sending  a mission demand for attending a conference for an administrative approval. He/She may exchange emails with multiple parties in his/her research lab or from the conference organizers. 
For these reasons, we consider emails separately without having any knowledge about their relations with each other.

In order to group emails into different clusters, we use the hierarchical clustering technique. Firstly, we choose a clustering and not a classification method, for the reason that the former is an unsupervised learning method. In our case, we have no knowledge about the dataset to be processed. Although each email may have some meta-data attributes describing it, there will be no information about the class of each email. Thus, we cannot apply a classification method, as we do not have a labeled training set. This justifies our use of the clustering technique.

Clustering is the grouping or segmenting of a set of objects, in our case the emails, into subsets or clusters, in which objects in one cluster are more related to one another than those of different clusters. There are two major methods for clustering: K-means and hierarchical clustering. 
Relative to the goal of this step, we choose the hierarchical clustering, as k-means clustering would have the following disadvantages in this context: 
\begin{itemize}
\item K-means is extremely sensitive to cluster center initialization.
\item Bad initialization can lead to bad convergence and bad overall clustering.
\item K-means requires as input the number of clusters, while we have no information about the number of clusters we need to obtain. 
\end{itemize}

For the above reasons, we choose to use an hierarchical clustering technique. More precisely, we use a "bottom-up" approach. Bottom-up algorithms treat each document as a singleton cluster.  Agglomeratively, the clusters are fused together according to a specific similarity (dissimilarity) measure between them. At each stage, the two clusters that are the most similar are joined together. There are several ways to define the similarity between two clusters in hierarchical clustering: (i) Single Linkage: where the distance between two clusters is the minimum distance between all pairs of emails ($e_{1}$,$e_{2}$) such that $e_{1}$ belongs to the first cluster and $e_{2}$ belongs to the second cluster; (ii) Complete Linkage: that is the opposite of Single Linkage, in which the distance between two clusters will be the maximum distance between the pairs of the emails of these clusters; (iii) Average Linkage: in which the distance between two clusters is defined as the average of all distances between the pairs of emails of these clusters. Experimentally, we find out that the complete linkage hierarchical clustering gives us the best clustering results. The reason behind this may be the suitability to consider the largest distance between the emails in the pairs of clusters so we can avoid clustering unintended clusters.

The challenge here is formulating the distance function that best serves our goal. The quality of the clustering highly depends on the distance function used. 
The distance function applied is as follows:\\\\
Dist($email_{1}$,$email_{2}$)= $w_{1}$.Eucl($subject_{1}$,$subject_{2}$) +$w_{2}$.Eucl($body_{1}$,$body_{2}$)\\\\
The distance between two emails (two rows) is calculated by taking into consideration the subject and the body of the message. However,  if we treat the subject and the body equally,  messages of the same subject will all be grouped in one cluster. To avoid this situation and ensure that the subject attribute will not have a dominating effect on the distance value, we provide a higher weight to the body part of the distance function $w2$ \textgreater $w1$. The values of $w1$ and $w2$ are obtained using experimental sensitivity analysis.

Actually, our topic-oriented email clustering method highly focuses on the email's body rather than its subject. Our assumption is that in some cases, users may exchange emails in the same thread under the same subject phrase in which the content of the message drifts to another topic. The email subject may be confusing sometimes because the email content may not conform with it, while the message body always reflects the real goal behind sending the email.
We conclude that in order to efficiently cluster emails according to their topic , we need focus on the message's body, i.e provide it a higher weight.

We apply the hierarchical clustering algorithm according to our customized distance function between the rows of the matrix obtained from the preprocessing phase including the subject of each email. The resulting clusters will contain all emails tackling a specific process model topic such as scheduling meeting or conference organization.
\section{Phase 3: Discovering the Process Model}
\label{sec:phase3}
\subsection{Finding the process instances}
After grouping the emails into different clusters according to the process they belong to, the next step aims at discovering process instances. For example, if a cluster contains all the messages exchanged for professional trips organization, we want to group all the message dealing with a specific trip organization (participation at COOPIS'16 conference), that is a process instance or case where the general process topic is the participation in a conference.

For this aim, we use again hierarchical clustering as emails grouping technique for the same reasons as in section \ref{sec:phase2}. The difference this time will be in the similarity formula. We calculate the distance between emails in one cluster by applying several combinations of attributes. We study the use of the combination of the three attributes subject, body and timestamp in the similarity estimation and their effects on the quality of process instances discovery. We make a comparison between sub-clustering quality when using the (i) body content, (ii) body and subject contents, or (iii) body, subject and timestamp contents in the distance measurement used for the hierarchical clustering algorithm.

Our experimental analysis (see section \ref{sec:usecase}) indicates the importance of time attribute in determining process instances rather than only considering the email content. 
We suppose that the emails belonging to the same process instance are likely to have relatively close timestamps. 
\subsection{Clustering emails of the same activity}
In the previous steps, we obtained  clusters of emails related to different process models, partitioned into sub-clusters corresponding to process instances. In this step, we aim to discover the activity applied in each email. In other words, we want to apply activity recognition techniques to associate emails with activities. Obtaining activities is one step forward towards discovering the business process model followed in each cluster and useful in itself for different process oriented analysis questions (What are the most frequent activities? What are the most time-inefficient activities ?) We do that by applying the k-means clustering algorithm. In the following we will explain our approach for activity recognition in emails and we justify the choice of k-means.

The input of this step will be a cluster which contains a set of process instances related to the same topic say: "meeting scheduling" (an element of the set of clusters obtained in the first step). Although process instances in the same cluster are executed to achieve the same goal, different instances may include different number of emails. The reasons behind that may be the loops that can happen to achieve a specific activity in the process instance such as "meeting proposal" or "meeting confirmation" and alternative paths or  optional activities executed only in some of the instances. 
K-means \cite{Kanungo-2002} requires as an input the number of clusters and its result may depend on the choice of the inital clusters. In the following we explain the customization of the k-means algorithm for our purpose.

\subsubsection{Choosing the number of clusters}
The k-means clustering algorithm is applied in order to obtain a  cluster for each specific activity in the process model. As the number of activities in a process model is not known, we estimate this number based on the number of emails in the process instances.
As instances of the same process model contains different number of activities (because of loops and alternative paths), we calculate the average number $N$ of emails contained in the process instances.

We choose the intial value for the number of clusters, $k$, equal to the estimated average number of activities ($k=N$). 
If we choose k\textless N, we may have emails that belong to different activities that are grouped in one cluster, and if we choose k\textgreater N, we may have more than one cluster containing emails representing the same activity. (The maximum number of activities instances in process instances may be greater than the number of activity types because of loops).

\subsubsection{Initializing the centroids}
Different intialization methods exist for assigning the initial centroids for the k-means algorithm.
In our case, the clusters will represent activity types, and the algortihm should partition the activities instances of each process instances in these clusters. Thus, emails in each process instance have to be dispatched in the $k$ clusters. 
For this reason, we choose a process instance containing a number of emails near the calculated average $N$  to intialize the clusters.
We then consider each email of this process instance as a centroid of a k-means cluster.
\subsubsection{Defining the distance} In order to get an efficient clustering result, we need to ensure that the distance measurement serves our needs. Our goal is to collect emails from different instances (of the same clsuter/process model) into groups each representing a specific activity. That's why, it is not sufficient to use the subject and the body of the message. The difference here is that we are not only looking after the global similarity between emails. We need to go deep into the email's details and meaning to deduce the activity to be accomplished in it. 
The distance calculation formula takes into consideration the meanings similarity of the words present in subject and body.  We use WordNet \cite{wordnet} to process the body term matrix in order to take into account synonyms. 
In future works, we plan to use the word2vec \cite{Mikolov-2013} which takes into consideration the linguistic contexts of words. 
Word2vec model \cite{Rong-2014}  provides the vector representations of words which carry semantic meanings and can be used to more accurately estimate the similarity of subject bag of words.
We should note that the time attribute will be excluded here as it has no effect on the clustering according to the activity. On another hand, the recipient of the message or the sender may be used for clustering, as activity types are usually assigned to the same role (for example, all the emails for booking tickets will have as a recipient the travel agent, while emails for booking rooms for  meetings will be addressed to the administrative person in charge with it). 

\subsubsection{Applying k-means}
Having defined the number of clusters and the distance measurement formula, we can apply the k-means clustering algorithm. The result of this step will be a set of groups, where each group contains a bunch of emails representing a specific activity. Emails in one group are sourced from different process instances. In some cases, we may have multiple mails that belong to the same process instance and are clustered in the same group. This indicates that multiple emails were exchanged to achieve the same activity. For instance, this may happen in the activity "meeting proposal" when the different parties do not agree on the same meeting date and time. So, several emails will be exchanged in this case to find the best timing that suits all parties. 
The k-means algorithm will be applied with different values for $k$, where $k$ varies around the average value $N$ obtained in the previous steps. The value of $k$ is choosen to the one that corresponds to the the best quality of clustering (in terms of clustering quality criteria such as purity, F-measure, or Rand index) or can be indicating by user after clusters inspection.

\subsection{Labeling emails with activity names} In this step, we will explain how we label the activities i.e. how to give activity name to all emails in one group (group obtained by the k-means clustering in the previous step) with the help of the user. 
After the end of the k-means clustering phase, the user is asked to label the medoid of each cluster. The given label will be considered the activity name of all emails present in that group (emails that are originating from different process instances). 
We can exploit the fact that each set of emails is labeled by an activity name to solve the problem researchers face when applying classification techniques for speech act recognition. As mentioned before, it is not practical to manually label all emails in the training set as an input for the classification algorithm. 
So these labels can be used to train activity recognition classifiers. More precisely, we plan to use the 1-nearest neighbor classifier on the cluster centers obtained by k-means to classify new email into the existing clusters.\\
The goal is to to predict for a new received email, the activity that it belongs to. We can exploit this to provide the user with recommendations about adding activities in his/her task list. 

\subsection{Deducing the process model}
\label{sec:step5}
Reaching this phase, each email is assigned to a cluster  (with clusterID) where a cluster represents a process topic, and to a process instance (with instanceID) and labeled with an activity name. These data can be used as an input for any process mining technique to obtain the business process model. In the log we obtained, each activity should be accompanied by its corresponding event. However for the moment we have supposed that the event is "complete". The goal of workflow mining is to extract information about processes from the input log. User can apply different process mining algorithms. Multiple algorithms are mentioned in the survey of Van der Aalst et al.\cite{Aalst-2004}. In Claes et al. \cite{Claes-2012} exposes the 5 most popular used process mining technique in ProM. The result of this step will be a process model for each cluster. This model will illustrate the process followed in the process instance (the email exchanges for achieving a specific goal).

\section{Implementation and Use Case Study}
\label{sec:usecase}
In this section, we present the current status of the implementation of the approach, illustrated by a usecase about a Ph.D. student emails. The emails extracted from the professional mailbox concern mission demand applications (attending a conference or a summer school) or meeting scheduling between several entities (Ph.D student and the responsible people in the research lab or the supervisors).

Figure \ref{fig:missiondemand} describes the steps of the process followed in an institution by a student who wants to apply for a conference mission or a summer school grant. The tasks are executed by email exchanges.  

We implemented the steps of the approach described in Figures  \ref{fig:approachSteps} and \ref{fig:processmodeling}, that take as input an email logs and produces a process log to be used by a process mining algorithm as R scripts.

We start by choosing a folder of emails taken from a student's email system. These emails do not provide any information about the relation between them (the threading relation or "In Reply to" relation). Surely, the chosen set  will include exchanged emails of different process topics. In other words, each bunch of emails in this set has been exchanged for applying a specific process in order to achieve an intended goal. Table 1. shows a screenshot of csv file containing the emails taken from the student's email log.

\begin{table}[htbp]
\begin{center}
\scriptsize{
\scalebox{0.57}{
\begin{tabular}{cccccc}
\hline
EmailID&Sender&Receiver&Subject&Timestamp&Body\\
\hline
1& diana.jlailaty@gmail.com & missionjc@dauphine.fr & mission demand & 2016-04-19 09:51:00 & Please find enclosed my\\&&&&& mission application for the Summer School\\&&&&&to be held in Urrugne from 5 to 10 June 2016\\
 \hline
2&missionjc@dauphine.fr & diana.jlailaty@gmail.com & mission demand & 2016-04-20 11:02:00 & 
I note that you have not yet linked\\&&&&& to the web page LAMSADE site. \\&&&&&Your mission will be taken into account.\\&&&&&Thank you to arrange and contact\\&&&&& us again as soon as possible.\\
 \hline
3&diana.jlailaty@gmail.com & missionjc@dauphine.fr & mission demand& 2016-04-20 02:35:00 & 
Thank you for considering my request.\\&&&&&
This is the link to my web page LAMSADE:\\&&&&& http://lamsade.dauphine.fr/~djlailaty/\\&&&&&
I am available for any further information.\\
\hline
4&missionjc@dauphine.fr &diana.jlailaty@gmail.com & mission demand & 2016-04-20 03:08:00 & 
Thank you Diana,\\&&&&&There is an error in the web address,\\&&&&& add to that In early WWW find your page.\\&&&&&Mission signed the request is attached to this email.\\&&&&&I invite you to visit the secretariat \\&&&&& eleni with the mission request.\\&&&&& You have been granted the amount of 550 euros.\\
\hline
5&diana.jlailaty@gmail.com &daniela.grigori@.dauphine.fr & meeting & 2016-03-29 10:34:00 & 
\\&&&&&What time is the meeting today?\\&&&&&\\
\hline
6&kbelhajj@googlemail.com &diana.jlailaty@gmail.com & meeting & 2016-03-29 10:42:000 & 
\\&&&&&Is it scheduled for 11am in Daniela office.\\&&&&&\\
\hline
7&diana.jlailaty@gmail.com &daniela.grigori@dauphine.fr & postpone the meeting & 2016-03-29 10:47:00 & 
Can the meeting today be on \\&&&&& 2:00 pm instead of 1:30 pm?\\&&&&& Because I want to attend \\&&&&& phd defense and pot of my colleague.\\
\hline
8&daniela.grigori@dauphine.fr &diana.jlailaty@gmail.com&postpone the meeting & 2016-03-29 10:52:00 & 
\\&&&&&For me, having the meeting in this time is Ok.\\
\hline
9&kbelhajj@googlemail.com &diana.jlailaty@gmail.com&postpone the meeting &2016-03-29 10:57:00& 
\\&&&&&It is good for me also.\\&&&&&\\
\hline
10&diana.jlailaty@gmail.com&daniela.grigori@dauphine.fr&set a meeting&2016-05-03 14:22:00& 
When will be our next meeting?\\&&&&& I am available the whole week.\\
\hline
13&diana.jlailaty@gmail.com&daniela.grigori@dauphine.fr&set a meeting&2016-05-03 14:22:00& 
When will be our next meeting?\\&&&&& I am available the whole week.\\
\hline
14&daniela.grigori@dauphine.fr&diana.jlailaty@gmail.com&set a meeting&2016-05-03 14:50:00& 
 I am available tomorrow for the meeting,\\&&&&& wednesday and friday.\\
\hline
15&kbelhajj@googlemail.com&daniela.grigori@dauphine.fr&set a meeting&2016-05-03 16:43:00& \\&&&&& Is the meeting tomorrow 10h good for both of you?\\&&&&&\\
\hline
16&kbelhajj@googlemail.com&diana.jlailaty@gmail.com&set a meeting&2016-04-22 17:50:00&  What have you done with the mission application?\\&&&&& Did you visit the secretariat Eleni for that?\\&&&&& You should do this as soon as possible.\\
\hline

20&diana.jlailaty@gmail.com&missionjc@dauphine.fr&mission demand&2016-06-25 10:35:00&  Please find attached my mission demand\\&&&&& to grenoble conference and summer school.\\
\hline
21&missionjc@dauphine.fr&diana.jlailaty@gmail.com&mission demand&2016-06-26 11:20:00& Your mission will be taken into account.\\&&&&& But before we confirm this,\\&&&&& you need to specify the detailed costs.\\&&&&& Thank you to arrange and\\&&&&& contact us again as soon as possible.\\
\hline
22&diana.jlailaty@gmail.com&missionjc@dauphine.fr&mission demand&2016-06-27 14:05:00&  Thank you for considering my demand.\\&&&&& Please find attached the detailed cost of my trip.\\
\hline
23&missionjc@dauphine.fr&diana.jlailaty@gmail.com&mission demand&22016-06-28 15:01:00&  Mission signed the request is attached to this email.\\&&&&& I invite you to visit the secretariat eleni\\&&&&& with the mission request. You have been granted\\&&&&& the amount of 600 euros.\\
\hline
\end{tabular}
}
}
\end{center}
\caption{Randomly chosen emails from student's emails log}
\end{table}
The table includes emails of different topics such as scheduling a meeting or mission demand application. As we've mentioned before, these emails are treated individually without taking into consideration any information about their threading relations.

Please note that email 16 illustrates the problem of subject drifting in reply emails while the same subject title is preserved (that was discussed in section \ref{sec:phase2}). This email has as subject title "set a meeting" while the content of the body is about the mission demand application discussed in emails 1 to 4. 

\begin{figure}[ht]
 \includegraphics[width=\linewidth]{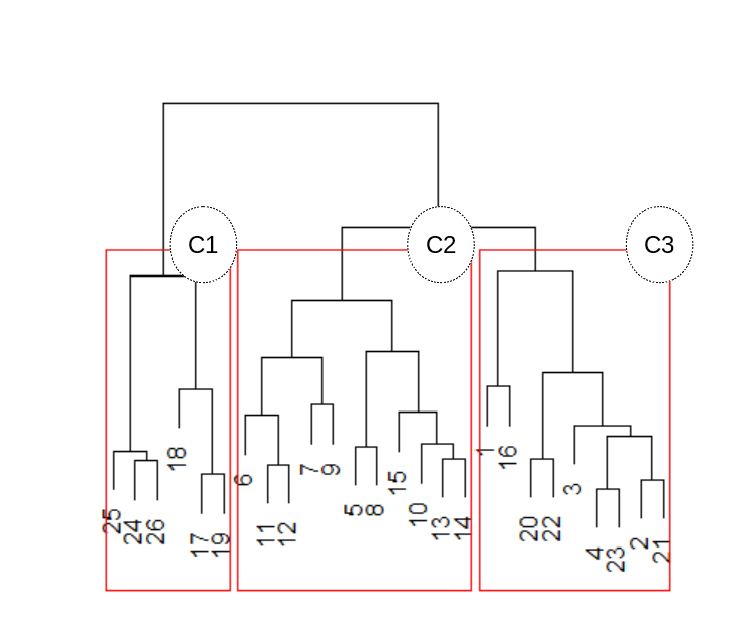}
  \caption{Clustered emails according to the topic (dendrogram plotted by R)}
  \label{fig:clusteringdendro}
\end{figure}

\begin{figure}[ht]
 \includegraphics[width=\linewidth]{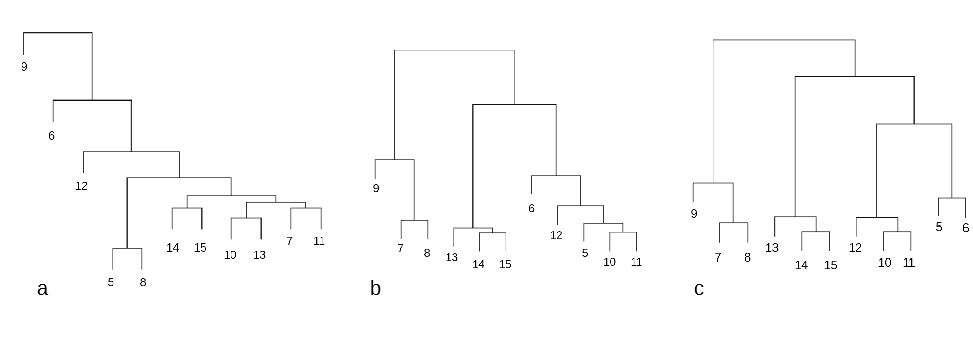}
  \caption{(a)Clustering according to body content, (b)Clustering according to body and subject content, (c)Clustering according to body, subject and timestamp content}
  \label{fig:clusteringabc}
\end{figure}

Process topic clustering will be applied on this set of emails. The R statistical programming language was used to implement the clustering algorithm. R language facilitates it with the huge number of built-in functions that can be used. R includes functions for hierarchical clustering. However, we implemented our customized clustering method with our customized distance function. The function calculating the distance between the emails takes as an input the normalized body term matrix containing all the words of all emails as columns headers and the email numbers as rows (emails are cleansed, matrix is normalized and processed as explained in section \ref{sub:normalization}). The function will then estimate the dissimilarity between two rows (two emails) and apply the hierarchical clustering accordingly. As mentioned in previous sections, the distance obtained for this clustering step focuses mainly on the email body text as we assume that it will hold all the information needed for collecting emails of the same process topic. Figure \ref{fig:clusteringdendro} shows how emails of the same topic are clustered together after hierarchically clustering the emails using our customized distance function.
\FloatBarrier
\FloatBarrier
Cluster C2 includes all emails with the context revolving around meeting scheduling. We can see that emails 5, 6, 7, 8, 9, 10, 11, 12, 13, 14 and 15 which all talk about meetings are clustered in one cluster. However, emails 1,2,3,4, and 16 which context is about a mission demand application are clustered together. We should realize that email 16 in fact belongs to a different different subject title than that of 1, 2, 3, and 4 but grouped in the same cluster. This indicates that our clustering algorithm mainly focuses on the message's content rather than other attributes.\\\\
After grouping emails according to their process model topic in a separate cluster, we will try now to find the process instances contained in each cluster. Again, we implemented an R function for hierarchical clustering. This time, we apply the clustering by trying three different distance functions with three different combinations of attributes. We cluster emails according to the email body, then according to the body and subject and at the end according to the body, subject and timestamp of each email. Each cluster obtained in the previous step will be input separately in this step. Let's consider cluster C2 focusing on meeting scheduling. Figure \ref{fig:clusteringabc} (a,b and c) shows us the results for the three types of clustering functions.
\FloatBarrier
\FloatBarrier 
If we go back to Table 1, we can see that emails from 5 to 15 can be divided into threads as following: (5,6), (7,8,9), (10,11,12), and (13,14,15). We realize that these threads are obtained as process instances in the second step. Figure \ref{fig:clusteringabc} (c). It is obvious (from Figure \ref{fig:clusteringabc} (b) ) how the clustering quality improves when we add the subject attribute (as the subject attribute gives good indication about the threading relation between emails regardless of some exceptions which we already tackled in the previous step). Nevertheless, the clustering quality improves more when we add the timestamp attribute which we claim that gives a good indication emails related to the same thread (emails in the same thread are likely to be sent in specific period of time). The same case happens for the process instances (1,2,3,4,16) and (20,21,21,23) in which each instance is considered for a separate mission demand.\\\\
We then work on collecting emails related to the same activity in one cluster. Again let us consider cluster C3 in Figure \ref{fig:clusteringdendro}. 
\begin{enumerate}\itemsep0em
\item Calculate the average number of emails in the instances of cluster C3. We get N is approximately equal to 4.
\item Choose an instance I containing 4 emails (user can help in choosing the instance). Let I=(20,21,22,23).
\item Initialize k=4 and the centroids of the clusters to each email in the chosen process instance I.
\item Apply the k-means clustering algorithm on all the emails in the cluster. The results will be as follows: clusters c1 ={1,20}, c2={2,21}, c3={3,22}, c4={4,23,16}.
\item We then ask the user to provide a label for each of the clusters. This provided label will be considered the activity or the task name of each of the emails contained in the labeled cluster. In our example, the user will label c1 (submit demand), c2 (request information), c3 (respond information), c4(accept demand or refuse demand).
\item In the next step, the log is enriched with activity names and given as input to an existing process mining tool. The resulting process oriented log can be used to analyse the underlying process in order to understand inefficient aspects and suggest improvements. For instance, we can discover that the loop in figure \ref{fig:missiondemand} between student and responsible is executed many times because some information is missing in student application. The process could be improved by providing the student with a standard form.

\FloatBarrier
\begin{figure}[ht]
 \includegraphics[width=\linewidth]{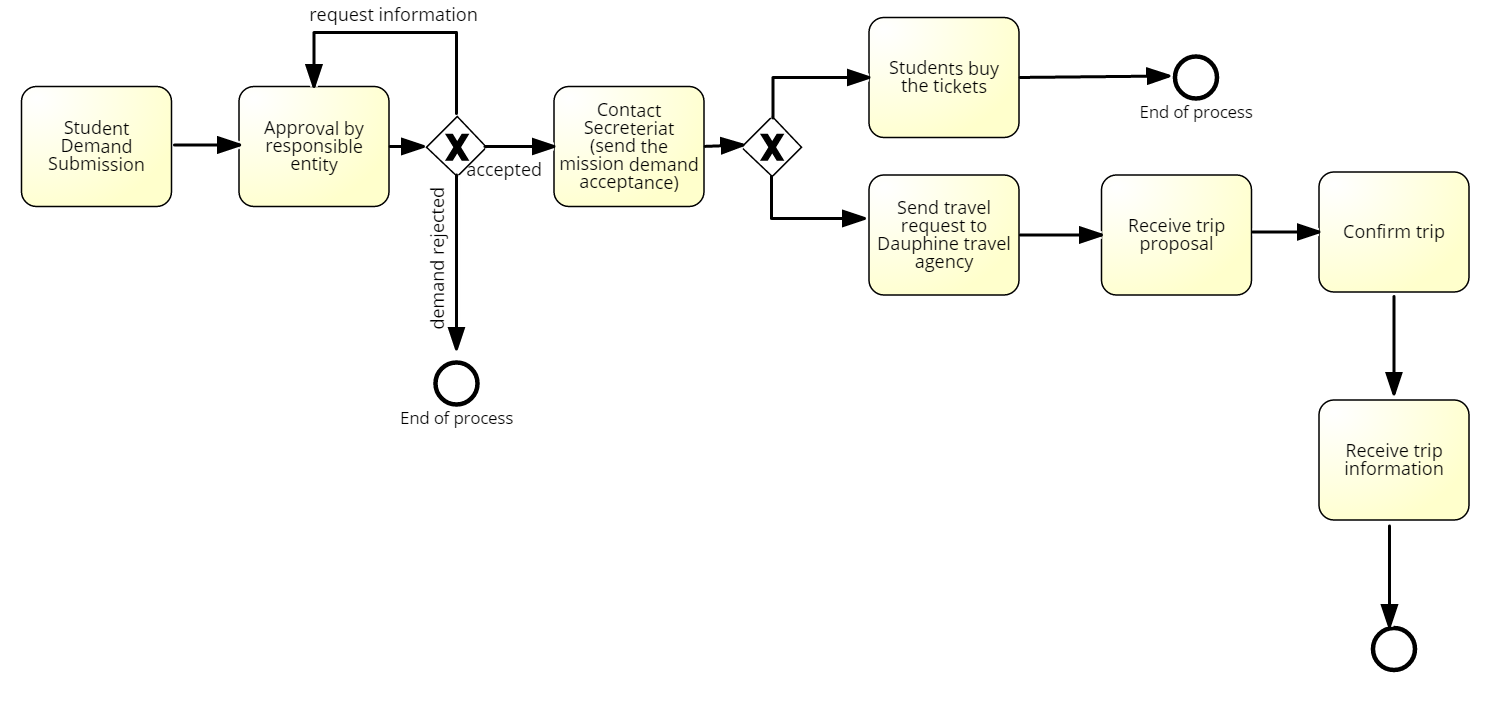}
  \caption{Process model for applying to a summer school or conference mission grant.}
  \label{fig:missiondemand}
\end{figure}
\FloatBarrier
\end{enumerate}
\section{Related Work}
\label{sec:relatedwork}
Until recently, there exist few works which combines the concepts of email analysis, activity recognition and business process discovery.

Similar to our work, the approach in Mavaddat et al. \cite{Mavaddat-2011} has been suggested to classify emails into business and non business oriented categories. The business-oriented emails are then grouped into threads using the similarity measurement to deduce process instances, from which the process model can be obtained fianlly. The authors suggest some preliminary ideas for labeling interactions between roles in process instances by using the classification of illocutionary speech acts assertive, directive, commissive, expressive, declarations suggested by Searl et al. \cite{Searl-1975}.

The approach applied by Khoussainov et al. \cite{Khoussainov-2006} exploits the relational structure between two problems: (i) extracting speech acts, (ii) finding related emails. Instead of attacking them separately, in their synergistic iterative approach, relations identification is used to assist semantic analysis, and vice versa.

SmartMail in Corston-Oliver et al. \cite{Corston-Oliver-2004} identifies action items(tasks) in email messages. It produces summary for each email which can be added by the user to his/her "to-do" list. In their approach, they need human annotators to provide tags to the training data set.\\
Despite the extensive studies of speech act recognition in many areas, developing speech act recognition for emails is very challenging. A major challenge is that emails usually have no labeled data for training statistical speech acts recognizers.

The work in Jeong et al. \cite{Jeong-2009} focuses on the problem of how to accurately recognize speech acts in emails by making maximum use of data from existing resources. They contribute in learning speech acts in a semi-supervised way by making use of some labeled data from spoken conversations. In their work, subtree features are exploited by using a subtree pattern mining. More precisely, they consider the text as a forest containing multiple trees. Each tree represents the relationships between several words (parent-child relationship). Dependency subtrees (speech acts) are then extracted from the trees forest. This is done by calling the models trained on data from existing external corpora and making use of it to extract the speech acts from the available emails.

On the other hand, there exist several works that uses fully supervised classification techniques in extracting speech acts. As an example, the authors of Qadir et al. \cite{Qadir-2011} aim to create sentence classifiers that can identify whether a sentence contains a speech act and can recognize sentences containing four different speech act classes: Commissive, Directive, Expressive, and Representative. They train classifiers to identify speech acts sentences using a variety of lexical, syntactic, and semantic features. The lexical and syntactic features are domain-independent while the syntactic feature is domain-dependent. To create their classifiers, they use the Weka \cite{Hall-2009} machine learning toolkit. They used Support Vector Machines (SVMs) with a polynomial kernel and the default settings supplied by Weka.

The authors of \cite{Cohen-2005} used text classification methods to detect “email speech acts”. Based on the ideas from Speech Act Theory \cite{Searl-1975} and guided by analysis of several email corpora, they defined a set of “email acts” (e.g., Request, Deliver, Propose, Commit) and then classified emails as containing or not a specific act. They showed that machine learning algorithms can learn the proposed email-act categories reasonably well. It was also shown that there is an acceptable level of human agreement over the categories.

Another category of works deal with the problem of conversation detection in email systems (see for exemple \cite{EreraC08}). While conversation detection is closed to the problem of process instance discovery, they are also different: a conversation is defined as taking place among the same group of people, while a process instance involves different persons, each one having a limited view on the overall set of exchanged emails (i.e., the travel agent only books the tickets, is not implied in the other exchanges).

 In Aalst et al. \cite{Aalst1-2007}, the authors develop the tool EmailAnalyzer which analyzes and transforms email messages in MS Outlook to a format that can be used by process mining tools. After a pre-porcessing step that disambiguates the receiver names and removes irrelevant messages,the e-mail log can be used to perform social network analysis (Sociograms and Messages Frequency Charts) and to generate the process log. In order to translate e-mail logs into process logs, the tool requires  that the message's subject contains tags representing the names of the case and of the activity. Contrary to this work that assumes that the task name is included into the message subject, our approach aims at discovering activity names and instances related to each message.
 
To conclude, we can see that many works addressed the problem of email classification, email speech act labeling, but only a few works combine the concepts of email analysis, activity recognition and business process discovery.

In email labels extraction in the related works, classification is used and classifiers are trained with a manually trained dataset (emails). In our work, we automatically obtain a labeled data set for future classification (for assigning activity names to incoming mails) by making use of the labeled clusters containing related set of emails. Regarding the grouping of emails into clusters of similar topics, contrary to other approaches that mostly use classification or k-means, hierarchical clustering is used in our approach as we suppose that we have no knowledge about the number of process topics tackled in the dataset. Compared to the work that is most similar to ours \cite{Aalst1-2007}, the approach presented in this paper leverages unsupervised machine learning techniques with little human involvement, while EmailAnalyzer supposes that the task name is included in the message subject.

\section{Conclusion}\label{sec:conclusions}
In this paper we proposed a framework for mining the process models underlying a log of emails. 
The approach is able to deal with emails exhanged in different process models. Hierarchical and k-means clustering algorithms are used to cluster emails according to the topic, then according to the process instance they belong to and finally for email activity labeling. We demonstrated a usecase that proves the usefulness of the proposed solution using logs containing emails of two different process topics: meeting scheduling and requests for conference travel grants.

In the future work, we plan to implement the activity recognition phase, using the clusters we obtained. We will also investigate how speech act approaches can help in more precisely defining the activity event associated to an email and activity loops. Moreover, we will extend the framework by integrating emails not only from one entity (in this paper, we considered as an example a Ph.D student), but also from several entities to cope with the whole process model (activities executed by all the parties).

We will also test the efficiency and the quality of or approach on a larger dataset. Fianlly, we will implement the recommendation system for incoming emails to recommend activities that can be added to the user's to-do list.

\bibliographystyle{plain}
\bibliography{bibliography.bib}
\end{document}